\documentclass[11pt]{article}
\usepackage{graphicx}
\usepackage{color}
\usepackage{amsmath}
\usepackage{amssymb}
\usepackage{amscd}
\usepackage{bbm}
\newcommand{\R}{\mathbb{R}}
\newcommand{\inr}[1]{\bigl< #1 \bigr>}

\newcommand{\E}{\mathbb{E}}

\newcommand{\eps}{\varepsilon}

\newcommand{\cF}{{\cal F}}
\newcommand{\cL}{{\cal L}}
\newcommand{\cD}{{\cal D}}
\newcommand{\cH}{{\cal H}}

\newtheorem{Theorem}{Theorem}[section]
\newtheorem{Lemma}[Theorem]{Lemma}

\newtheorem{Definition}[Theorem]{Definition}
\newtheorem{Problem}[Theorem]{Problem}

\newtheorem{Corollary}[Theorem]{Corollary}

\newtheorem{Assumption}{Assumption}[section]

\numberwithin{equation}{section}

\def \proof {\noindent {\bf Proof.}\ \ }

\def \endproof
{{\mbox{}\nolinebreak\hfill\rule{2mm}{2mm}\par\medbreak}}
\def\IND{\mathbbm{1}}

\begin{document}
\title{\bf Learning without Concentration
}

\author{
\\{\bf Shahar Mendelson}
\thanks{Department of Mathematics, Technion -- Israel Institute of Technology, email: shahar@tx.technion.ac.il. Partially supported by the Mathematical Sciences Institute -- The Australian National University and by ISF grant 900/10.}
}

\maketitle

\begin{abstract}
We obtain sharp bounds on the performance of Empirical Risk Minimization performed in a convex class and with respect to the squared loss, without assuming that class members and the target are bounded functions or have rapidly decaying tails.

 Rather than resorting to a concentration-based argument, the method used here relies on a `small-ball' assumption and thus holds for classes consisting of heavy-tailed functions and for heavy-tailed targets.

 The resulting estimates scale correctly with the `noise level' of the problem, and when applied to the classical, bounded scenario, always improve the known bounds.
\end{abstract}

\section{Introduction} \label{sec:intro}
Our aim is to study the error of Empirical Risk Minimization (ERM), performed in a convex class and relative to the squared loss.

To be more precise, let $\cF$ be a class of real-valued functions on a probability space $(\Omega,\mu)$ and let $Y$ be an unknown target function. One would like to find some function in $\cF$ that is almost the `closest' to $Y$ in some sense.

A rather standard way of measuring how close $Y$ is to $\cF$, is by using the squared loss $\ell(t)=t^2$ to capture the `point-wise distance' $(f(x)-y)^2$, and being `close' is measured by averaging that point-wise distance. Hence, if $X$ is distributed according to the underlying measure $\mu$, the goal is to identify, or at least approximate with good accuracy, the function $f^* \in \cF$ that minimizes $\E(f(X)-Y)^2=\|f(X)-Y\|_{L_2}^2$ in $\cF$, assuming, of course, that such a minimizer exists.

Unlike questions in Approximation Theory, the point in learning problems is to approximate $f^*$ using random data -- an independent sample $(X_i,Y_i)_{i=1}^N$ selected according to the joint distribution defined by $\mu$ and the target $Y$.

A more `statistical' way of describing this problem is the minimization of the average cost of a mistake. If the price of predicting $f(X)$ instead of $Y$ is $(f(X)-Y)^2$, the average cost is $\E(f(X)-Y)^2=\|f(X)-Y\|_{L_2}^2$. Hence, one would like to approximate the minimizer in $\cF$ of the average cost, but with only the partial information of the given random sample at one's disposal.

\vskip0.5cm

It should be noted that approximating $f^*$ using random data (the so-called {\it estimation problem}), is just one of the two natural questions in this context. The other, called the {\it prediction problem}, deals with identifying a function in $\cF$ whose `predictive capabilities', reflected by $\E(f(X)-Y)^2$, are almost the same as the best possible in the class, $\E(f^*(X)-Y)^2$. To simplify the exposition, we will focus on the estimation problem relative to the squared loss, and refer the reader to \cite{Men-extend-ver} for the study of both prediction and estimation relative to a general convex loss function.

\vskip0.5cm

One way of using the given data $(X_i,Y_i)_{i=1}^N$ is by selecting a random element in $\cF$, denoted by $\hat{f}$, that minimizes the empirical loss
$$
P_N \ell_f = \frac{1}{N} \sum_{i=1}^N (f(X_i)-Y_i)^2,
$$
where here, and throughout this article, $P_N g $ denotes the empirical mean of the function $g$ with respect to the given sample.

With this choice of a learning procedure, $\hat{f}$ is called the {\it empirical minimizer}, and the procedure that selects $\hat{f}$ is Empirical Risk Minimization (ERM).

In the context of the estimation problem, it is natural to measure the success of ERM and the effectiveness of the choice of $\hat{f}$ in the following way: that for `most' samples $(X_i,Y_i)_{i=1}^N$, ERM produces a function that is close in $L_2(\mu)$ to the best approximation of the target $Y$ in $\cF$; that is, a high probability upper estimate on
$$
\|\hat{f}-f^*\|_{L_2}^2=\int_\Omega |\hat{f}(x)-f^*(x)|^2d\mu(x)
$$
for the empirical minimizer $\hat{f}$ selected according to the data $(X_i,Y_i)_{i=1}^N$.

\vskip0.5cm

Our starting point is a well known result that deals with this very question: controlling the distance in $L_2(\mu)$  between the function produced by ERM and $f^*$. Theorem \ref{thm:Kolt}, formulated below, has been established in \cite{BBM} (see Corollary~5.3 there and also Theorem~5.1 in the survey \cite{Kolt}).

Let $\cD_{f^*}$ be the $L_2(\mu)$ ball of radius $1$, centred in $f^*$. Thus,
$\{f \in \cF : \|f-f^*\|_{L_2} \leq r\} = \cF \cap r\cD_{f^*}$.
For any $r>0$, let
\begin{equation} \label{eq:k-N}
k_N(r)=\E\sup_{f \in \cF \cap 2\cD_{f^*}} \left|\frac{1}{\sqrt{N}}\sum_{i=1}^N\eps_i (f-f^*)(X_i)\right|,
\end{equation}
where $(\eps_i)_{i=1}^N$ are independent, symmetric, $\{-1,1\}$-valued random variables (random signs) that are independent of $(X_i)_{i=1}^N$, and the expectation is with respect to both $(\eps_i)_{i=1}^N$ and $(X_i)_{i=1}^N$.

Set
$$
k_N^*(\gamma)=\inf \left\{r>0: k_N(r)\leq \gamma r^2\sqrt{N}\right\}.
$$
\begin{Theorem} \label{thm:Kolt}
There exist absolute constants $c_0,c_1$ and $c_2$ for which the following holds.
If $\cF \subset L_2(\mu)$ is a closed, convex class of functions that are bounded by $1$ and the target $Y$ is also bounded by $1$, then for every $t>0$, with probability at least $1-c_0\exp(-t)$,
\begin{equation}
  \label{eq:koltchinskii}
\|\hat{f}-f^*\|_{L_2}^2 \leq c_1\max\left\{\left(k_N^*(c_2)\right)^2,\frac{t}{N}\right\}.
\end{equation}
\end{Theorem}

\subsection{The many downsides of Theorem \ref{thm:Kolt}} \label{sec:downside}
 The proof of Theorem \ref{thm:Kolt} relies heavily on the fact that $\cF$ consists of functions that are bounded by $1$ and that the target is also bounded by $1$. Both are restrictive assumptions and exclude many natural problems that one would like to consider.
\begin{description}
\item{1.} {\bf Gaussian noise:} arguably the most basic statistical problem is when $Y=f_0(X)+W$ for some $f_0 \in \cF$ and $W$ that is a centred gaussian variable with variance $\sigma$ that is independent of $X$. Thus, the given data consists of `noisy' measurements of $f_0$ corrupted by gaussian noise.

    Since a gaussian random variable is unbounded, Theorem \ref{thm:Kolt} cannot be used to address the estimation problem that involves gaussian noise, regardless of the choice of $\cF$.
\item{2.} {\bf Heavy-tailed noise:} The vague term `heavy-tailed function' is used to describe a function for which the measure of its tail, $Pr(|f|>t)$, decays to zero relatively slowly -- for example, polynomially in $1/t$. In particular, such a function need not be bounded. Hence, any kind of an estimation problem that involves a heavy-tailed target $Y$ cannot be treated using Theorem \ref{thm:Kolt}.

\item{3.} {\bf Gaussian regression:} Let $T \subset \R^n$ and set $\cF=\left\{\inr{t,\cdot} : t \in T\right\}$ to be the class of linear functionals indexed by $T$. If the underlying measure $\mu$ is the standard gaussian measure on $\R^n$, then for every $t \in T$, $f_t(X)=\inr{t,X}$ is unbounded. Thus, regardless of the target, it is impossible to apply Theorem \ref{thm:Kolt} to a problem that involves the class $\cF$.

\item{4.} {\bf General regression:} A class of linear functionals on $\R^n$ is not  bounded almost surely unless the underlying measure $\mu$ has a compact support. What is rather striking is that even in problems in which $\mu$ does have a compact support and which seemingly belong to the bounded framework, Theorem \ref{thm:Kolt} is far from optimal.
\end{description}

To give a rather natural and well studied example that fits the bounded framework but for which the outcome of Theorem \ref{thm:Kolt} is far from optimal, let $B_1^n=\left\{x \in \R^n : \sum_{i=1}^n |x_i| \leq 1\right\}$ be the unit ball in $\ell_1^n$, set $T_R=R B_1^n$ for some $R>0$ and put $\cF_R=\left\{\inr{t,\cdot}: \ t \in T_R\right\}$. Estimation problems in $\cF_R$ have been of particular interest in recent years, mainly because of their obvious connections to sparse recovery procedures like compressed sensing or LASSO.

Let $X=(\eps_i)_{i=1}^n$ be a random vector whose coordinates are independent random signs (i.e. symmetric, $\{-1,1\}$-valued random variables). Observe that $X$ is an isotropic random vector, since its covariance structure coincides with the standard Euclidean structure on $\R^n$. Indeed, for every $t \in \R^n$, $\E \inr{X,t}^2=\|t\|_{\ell_2^n}^2$, where $\|t\|_{\ell_2^n}$ is the standard Euclidean norm on $\R^n$.

Let $\eps_{n+1}$ be a random sign that is independent of $(\eps_i)_{i=1}^n$, fix $0 \leq \sigma \leq R$ and $t_0 \in T_R$, and put $Y=\inr{t_0,\cdot}+\sigma \eps_{n+1}$.

Note that for such a target $Y$, $f^*(X)=\inr{t_0,X}$, and that the estimation problem of $Y$ in $\cF_R$ belongs to the bounded framework: for every $t \in T_r$,
$\|\inr{t,X}\|_{L_\infty} \leq \max_{1 \leq i \leq n} |\eps_i| \cdot \sum_{i=1}^n |t_i| \leq R$,
and $\|Y\|_{L_\infty} \leq R+\sigma$. However, as will be explained in Section \ref{sec:examples}, the estimate resulting from Theorem \ref{thm:Kolt} on $\|\hat{f}-f^*\|_{L_2}=\|\hat{t}-t_0\|_{\ell_2^n}$, is far from optimal, and scales incorrectly both with $R$ and with $\sigma$.

More accurately, we will show that the outcome of Theorem \ref{thm:Kolt} is as follows: there are absolute constants $c_1,c_2$ and $c_3$ and
$$
\rho_N=
\begin{cases}
\frac{R^2}{\sqrt{N}} \sqrt{\log \left(\frac{2c_1 n}{\sqrt{N}}\right)} & \mbox{if} \ \ N \leq c_1 n^2
\\
\\
\frac{R^2n}{N} & \mbox{if} \ \ N >c_1 n^2,
\end{cases}
$$
for which, with probability at least $1-2\exp(- c_2N \rho_N/R^2)$, ERM produces $\hat{t} \in RB_1^n$ that satisfies $\|\hat{t}-t_0\|_{\ell_2^n}^2 \leq c_3\rho_N$.

On the other hand, it follows from \cite{LM1} that the correct rate for this problem is very different: if
\begin{equation*}
v_1=
\begin{cases}
\frac{R^2}{N} \log\left(\frac{2c_4n}{N}\right) &  \mbox{if} \ \ N \leq c_4 n,
\\
\\
0 & \mbox{if} \ \ N > c_4 n
\end{cases}
\end{equation*}
and
\begin{equation*}
v_2 =
\begin{cases}
\frac{R\sigma}{\sqrt{N}} \sqrt{\log\left(\frac{2c_5n\sigma}{\sqrt{N}R}\right)} & \mbox{if } \ \  N \leq c_5n^2 \sigma^2/R^2
\\
\\
\frac{\sigma^2 n}{N} & \mbox{if} \ \ N >c_5n^2 \sigma^2/R^2,
\end{cases}
\end{equation*}
then with probability at least $1-2\exp\left(-c_6N \min\{v_2,1\}\right)$,
$$
\|\hat{t}-t_0\|_{\ell_2^n}^2 \leq c_7\max\left\{ v_1,v_2\right\}
$$
for absolute constants $c_4,c_5,c_6$ and $c_7$.

Therefore, the outcome of Theorem \ref{thm:Kolt} scales incorrectly with $R$ and with $\sigma$, and most notably, the estimation error does not converge to zero when $\sigma$ becomes smaller and the problem is almost `noise-free'.

\vskip0.5cm

This example belongs to the {\it persistence framework}, which will be explored in greater detail later on. It, and other examples like it show that the poor outcome of Theorem \ref{thm:Kolt} is endemic rather than merely an accident.

To get a clearer picture of the reasons why Theorem \ref{thm:Kolt} is truly suboptimal, let us take a closer look at its assumptions.

\subsubsection*{Why boundedness?}
The point-wise boundedness assumption in Theorem \ref{thm:Kolt} has been used frequently in Learning Theory, and for a good reason. It allows one to invoke two important tools from empirical processes theory, which are simply not true in a more general setup.
\begin{description}
\item{1.} {\it Contraction methods}:  Since the interval $[-2,2]$ contains all possible values $\{f(X)-Y : f \in \cF\}$, and since the loss function $\ell(t)=t^2$ is a Lipschitz function with a well behaved constant on that interval, one may use  contraction arguments, that, roughly put, allow one to control the supremum of empirical process $f \to |P_N \ell_f - P\ell_f|$ using the process $f \to |P_N f -P f|$ (where we use the standard notation $P_N f \equiv \frac{1}{N}\sum_{i=1}^N f(X_i)$, $P f  \equiv \E f$ and $\ell_f(X,Y)=\ell(f(X)-Y)$).
\item{2.} {\it Concentration methods}:  Talagrand's concentration inequality for empirical processes indexed by a class of uniformly bounded functions \cite{Tal94,Led} has played an essential role in Learning Theory - and Theorem \ref{thm:Kolt} is no exception. It is used to show that with high probability, the supremum of an empirical process is almost the same as its expectation, as well as the {\it Rademacher averages} of the class (see, e.g., \cite{BBM,Kolt} and references therein for more details on the role of Rademacher averages in Learning Theory).
\end{description}

It takes no more than a glance to see that contraction and concentration are crucial to the proof of Theorem \ref{thm:Kolt}. A more careful examination shows that there is no realistic hope of extending the proof beyond the restricted setup of classes of uniformly bounded functions and a bounded target without introducing a totally new argument.

\subsubsection*{The noise barrier}
An important observation that reflects the suboptimal nature of Theorem \ref{thm:Kolt} is that the estimation error is insensitive to the {\it noise level}. The term `noise level' is used here to describe the distance between the target and the class, either with respect to the $L_2$ norm, or with respect to a different $L_p$ norm, depending on the situation\footnote{The reason for naming the distance between the class and the target `noise level' is the very simple choice of a target $Y=f_0(X)+W$ for some $f_0 \in \cF$, and an independent, centred random variable $W$, representing the noise. It is straightforward to verify that for such a target $f^*=f_0$, and the $L_2$ distance between the class and the target is $\|f^*(X)-Y\|_{L_2}=\|W\|_{L_2}$ -- the variance of $W$.}.

One would expect that when the noise level (distance) tends to zero and the problem becomes almost noise-free (or {\it realizable}), $\|\hat{f}-f^*\|_{L_2}$ should decrease, and in some cases should even tend to zero. Among these problems are compressed sensing and phase recovery, in which the exact recovery of the signal is possible (see the discussion in \cite{LM1,LM2}). Unfortunately, Theorem \ref{thm:Kolt} is insensitive to the noise level, and this by-product of its proof cannot be resolved without totally changing the method of analysis.

\vskip0.5cm

All these observations are strong indications that if Theorem \ref{thm:Kolt} is to be radically improved, the entire concentration-contraction framework on which it is based, must be abandoned. Moreover, any alternative framework should address two core issues:
\begin{description}
\item{$\bullet$} It must be able to handle `heavy-tailed' problems.
\item{$\bullet$} The estimation error must scale correctly with the key parameters of the problem, especially with the noise level.
\end{description}

\section{Towards a heavy-tailed framework}
In view of the requirements outlined above, an improved framework has to contend with functions that may have heavy tails, and certainly need not be bounded. We will first explain why standard concentration-based arguments fail miserably when faced with a heavy-tailed scenario. We will then suggest an alternative to standard concentration, introduce the complexity parameters that arise naturally in the suggested framework, and explain their statistical interpretation.

\subsection{Bypassing concentration}
The title of this article may create the wrong impression -- that concentration methods are not needed and will not take any part in the analysis of ERM. Actually, the way the title should be understood is different: that learning is possible even when concentration is false.

As a starting point, and seeing that one is interested in the squared loss, one should note the substantial difference between a two-sided concentration inequality, stating, for example, that with high probability,
$$
\left|\frac{1}{N}\sum_{i=1}^Nf^2(X_i)-\E f^2\right| \leq \frac{1}{2} \E f^2,
$$
and just the lower bound on the empirical mean,
$$
\frac{1}{2} \E f^2 \leq \frac{1}{N}\sum_{i=1}^Nf^2(X_i).
$$
Even at a first glance, a significant difference between the two inequalities is not far-fetched: while one very large value of $|f(X_i)|$ may spoil the two-sided condition (because $P_N f^2$ will be larger than $(3/2)\E f^2$), it can only help the lower bound on $P_N f^2$.

As a concrete example, fix an integer $N \geq 100$ and let $Z$ be a random variable for which $Pr(|Z|=2\sqrt{N})=1/N^2$, and $|Z| = 1$ otherwise. It is straightforward to verify that $\E Z^2=1+4/N-1/N^2$ and that $\|Z\|_{L_4}/\|Z\|_{L_2} \leq 3$.

If $Z_1,...,Z_N$ are independent copies of $Z$, then with probability at least $1/2N$ there exists some $1 \leq i \leq N$ for which $|Z_i| = 2\sqrt{N}$. On that event,
$$
\frac{1}{N}\sum_{i=1}^N Z_i^2 \geq 4 \geq \frac{3}{2} \E Z^2,
$$
and thus
$$
Pr\left(\left|\sum_{i=1}^N Z_i^2 - \E Z^2\right| \leq \frac{1}{2}\E Z^2\right) \leq 1- \frac{1}{2N}.
$$
On the other hand, one may verify that
$$
Pr\left(\frac{1}{N} \sum_{i=1}^N Z_i^2 \geq \frac{1}{2} \E Z^2 \right) \geq 1- 2\exp(-c_1N)
$$
for a suitable absolute constant $c_1$ -- and that a similar estimate is true for any random variable $Z$ for which $\|Z\|_{L_4}/\|Z\|_{L_2}$ is well behaved, with $c_1$ depending only on this ratio.

\vskip0.5cm

The difference between an upper estimate and a lower one is even more obvious if one only assumes that $Pr(|f| \geq u) \geq \eps$ for fixed constants $u$ and $\eps$. Using a simple binomial estimate, one may show that with probability at least $1-2\exp(-c_2\eps N)$, $P_N f^2 \geq \eps u^2/2$ for a suitable absolute constant $c_2$. However, there is no hope of obtaining any upper estimate on $P_N f^2$ based on the given information.

An immediate consequence of these observations is that the standard method of analysis for the estimation problem, which is based on a two-sided concentration argument that holds with exponential probability, can never work in heavy-tailed situations. Thus, one must find a different argument altogether if one wishes to deal with learning problems that include classes of heavy-tailed functions or with a heavy-tailed target.

The difference between a two-sided estimate and a lower bound takes centre-stage when one realizes that the main component in solving the estimation problem is actually a {\it lower bound} on 
$$
\inf_{\{f \in \cF : \ \|f-f^*\|_{L_2} \geq r\}} \frac{1}{N} \sum_{i=1}^N \left(\frac{f-f^*}{\|f-f^*\|_{L_2}}\right)^2(X_i)
$$
rather than a two-sided bound. The significance of this lower bound will be made clear in Section \ref{sec:method}. It will eventually lead to a sharp estimate on $\|\hat{f}-f^*\|_{L_2}$ even when two-sided concentration is impossible.

\subsection{Two parameters for two regimes} \label{sec:regimes}
When one considers the performance of a learning procedure, it is reasonable to expect two different `performance regimes', according to the difficulties the learner faces. As will be explained below, there are clear differences between `low noise' problems, in which the target $Y$ is close to $\cF$ and the rate one should expect is close to the noise-free (realizable) rate, and `high noise' problems, when $Y$ is far enough from $\cF$ and the interaction between the target and the class determines the estimation error. One may also expect that just as in a realizable problem, the complexity parameter governing the `low-noise' regime is intrinsic to $\cF$ and therefore should not depend on $Y$ at all, while the parameter controlling the `high-noise' regime depends on $Y$ in one way or another.

\subsubsection*{Controlling the Version space - quadratic estimates}
We begin with the definition of the first parameter we will be interested in, and which governs the `low-noise' regime.
\begin{Definition} \label{def:beta}
Given a class of functions $\cF$ and $\gamma>0$, set
$$
\beta_N^*(\gamma) = \inf\left\{r>0 :  \E \sup_{f \in \cF \cap r \cD_{f^*}} \left|\frac{1}{\sqrt{N}}\sum_{i=1}^N \eps_i (f-f^*)(X_i) \right| \leq \gamma \sqrt{N}r \right\},
$$
where the expectation is taken with respect to both $(\eps_i)_{i=1}^N$ and $(X_i)_{i=1}^N$.
\end{Definition}
Note that $\beta_N^*$ is indeed an intrinsic parameter, in the sense that it depends only on the class $\cF$ and not on the exact nature of the `noise' $f^*(X)-Y$. From a purely technical perspective, $\beta_N^*$ measures when the Rademacher averages of the `localized' set $\{f-f^* : f \in \cF \cap r\cD_{f^*}\}$  scale like $r$ rather than like the normalization $r^2$, which has been used in the definition of $k_N^*$ and in Theorem \ref{thm:Kolt}. Thus, $\beta_N^*$ will always be much smaller than $k_N^*$ when dealing with $r \ll 1$, as we do.

Off-hand, the meaning and significance of $\beta_N^*$ is not obvious. It is, perhaps, surprising that it captures properties of the {\it version space} of the problem.

The version space is a random subset of $\cF$ that consists of all the functions in the class that agree with $f^*$ on the sample $(X_i)_{i=1}^N$. When the problem is noise-free ($Y=f^*$), a learning procedure makes significant mistakes only when there are functions in $\cF$ that, despite being `far-away' from $f^*$, still satisfy that $f(X_i)=f^*(X_i)$ for every $1 \leq i \leq N$. In other words, significant mistakes occur in a noise-free problem only when the version space is large.

As noted earlier, it seems plausible that when the noise level is low rather than zero, that is, when $Y$ is close enough to $\cF$, the situation does not change significantly and mistakes are essentially due to a large version space.
Thus, when trying to bound the error of ERM, the first order of business is to identify a parameter that captures the `size' of the version space, and $\beta_N^*$ gives thats and much more.

We will show that with high probability, if $\|f-f^*\|_{L_2} \geq \beta_N^*$, then
\begin{equation} \label{eq-lower-est-quad-intro}
\frac{1}{N}\sum_{i=1}^N (f-f^*)^2(X_i) \geq c\|f-f^*\|_{L_2}^2
\end{equation}
for an appropriate constant $c$, and with (almost) no assumption on $\cF$ (see Theorem \ref{thm:quad-est} for the exact formulation).

Immediate outcomes of \eqref{eq-lower-est-quad-intro} are that the version space cannot include functions for which $\|f^*-f\|_{L_2} \geq \beta_N^*$ and that sampling is `stable' for functions that are not too close to $f^*$. Indeed, \eqref{eq-lower-est-quad-intro} implies that on a large event, the empirical distances $(N^{-1}\sum_{i=1}^N (f-f^*)^2(X_i))^{1/2}$ are at least a fixed proportion of the original $L_2(\mu)$ distances $\|f-f^*\|_{L_2}$.

\subsubsection*{Controlling the interaction with the noise}

The second regime is encountered once the noise level increases, and mistakes happen for a totally different reason: the `interaction' of the target $Y$ with class members. It turns out that this interaction is captured by the following parameter.

Let $\xi(X,Y)=f^*(X)-Y$ represent the noise\footnote{calling $\xi$ `the noise' is a natural name when $Y=f_0(X)+W$ for some $f_0 \in \cF$ and $W$ that is independent of $X$. We will use that name even when $Y$ does not have that form.} and set
\begin{equation} \label{eq:phi}
\phi_N(s)= \sup_{f \in \cF \cap s\cD_{f^*}} \left|\frac{1}{\sqrt{N}} \sum_{i=1}^N \eps_i \xi_i (f-f^*)(X_i)\right|,
\end{equation}
where for every pair $(X_i,Y_i)$, $\xi_i=f^*(X_i)-Y_i$, and $(\eps_i)_{i=1}^N$ are, as always, random signs that are also independent of $(X_i,Y_i)_{i=1}^N$. Thus $\phi_N(s)$ is defined on the probability space $(\Omega \times \R \times \{-1,1\})^N$, relative to the product measure endowed by $(X_i,Y_i)_{i=1}^N$ and $(\eps_i)_{i=1}^N$.

Let
$$
\alpha_N^*(\gamma,\delta) = \inf\left\{ s>0 : Pr \left(  \phi_N(s)\leq \gamma s^2 \sqrt{N} \right) \geq 1-\delta \right\}.
$$
The fixed point $\alpha_N^*$ is of a similar nature to $k_N^*$ and the two share the same scaling, of the order of $\sqrt{N} s^2$, but with one key difference: the supremum of the multiplier process in \eqref{eq:phi} measures the maximal correlation elements of the random set $\{ (f-f^*)(X_i) : f \in \cF \cap s\cD_{f^*}\}$ have with the {\it random symmetrized noise vector} $(\eps_i \xi_i)_{i=1}^N$, while the random process used in the definition of $k_N^*$,
$$
\sup_{f \in \cF \cap s\cD_{f^*}}\left|\frac{1}{\sqrt{N}} \sum_{i=1}^N \eps_i (f-f^*)(X_i)\right|,
$$
measures the maximal correlation elements of the same random set have with the random vector $(\eps_i)_{i=1}^N$. The obvious difference is that $(\eps_i)_{i=1}^N$ represents a `generic' noise and has nothing to do with the specific noise that the learner has to deal with.

Note that if functions in $\cF$ and $Y$ happen to be bounded by $1$, then $\xi=f^*(X)-Y$ is bounded by $2$. Applying a standard contraction argument (see, e.g. \cite{LT,VW}) it follows that for every $u>0$,
\begin{equation} \label{eq:contraction}
Pr \left(\sup_{f \in \cF \cap s\cD_{f^*}}  \left|\sum_{i=1}^N \eps_i \xi_i (f-f^*)(X_i)\right| > u \right) \leq 2Pr \left(\sup_{f \in \cF \cap s\cD_{f^*}}  \left|\sum_{i=1}^N \eps_i (f-f^*)(X_i)\right| > \frac{u}{2} \right),
\end{equation}
and thus one may show that $\alpha_N^*$ is dominated by $k_N^*$ in the bounded case. Moreover, while $k_N^*$ is insensitive to the noise level, because the contraction argument used in \eqref{eq:contraction} destroys any dependence on the `noise multipliers' $(\xi_i)_{i=1}^N$, $\alpha_N^*$ is highly affected by the noise -- and in a favourable way. When $Y$ is close to $\cF$ in the right sense, $\xi$ is close to zero, leading to smaller multipliers $(\xi_i)_{i=1}^N$, and thus to a smaller fixed point $\alpha_N^*$.

\section{The main result} \label{sec:method}
Next, let us explain why this heuristic description of the roles of $\alpha_N^*$ and $\beta_N^*$ is reasonable, and why splitting the estimation problem to two components, each captured by one of the two parameters, is the first step in bypassing the concentration-contraction mechanism used in the proof of Theorem \ref{thm:Kolt}.

\vskip0.5cm
For every $f \in \cF$, let ${\cal L}_f$ be the excess loss functional associated with $f$, which is defined by
$$
\cL_{f}(X,Y)=\ell_f(X,Y)-\ell_{f^*}(X,Y)=(f(X)-Y)^2 - (f^*(X)-Y)^2.
$$
Therefore,
\begin{align} \label{eq:squared-loss}
\cL_f(X,Y) = & (f-f^*)^2(X) + 2(f-f^*)(X)(f^*(X)-Y) \nonumber
\\
= & (f-f^*)^2(X) + 2\xi(f-f^*)(X),
\end{align}
where, as always, $f^*$ is the unique minimizer in $\cF$ of $\E(f(X)-Y)^2$ and $\xi(X,Y)=f^*(X)-Y$.

Let $\cL_{\cF} = \{\cL_f : f \in \cF\}$ be the excess loss class and note that it has two important properties.
\begin{description}
\item{$\bullet$}  $\cL_{\cF}$ is a shift of the class $\{(f(X)-Y)^2 : f \in \cF\}$ by the fixed function $\ell_{f^*}=\xi^2$. Thus, an empirical minimizer of the loss is an empirical minimizer of the excess loss.

\item{$\bullet$} Since $0 \in \cL_{\cF}$, the empirical minimizer $\hat{f}$ satisfies $P_N \cL_{\hat{f}} \leq 0$.
\end{description}
Therefore, if $(X_1,...,X_N)$ is a sample for which
$$
\{f \in \cF : \|f-f^*\|_{L_2} \geq  \rho \} \subset \{f \in \cF : P_N \cL_f > 0\},
$$
then $\|\hat{f} - f^*\|_{L_2} < \rho$ -- simply because an empirical minimizer cannot belong to the set $\{f \in \cF : \|f-f^*\|_{L_2} \geq  \rho \}$.

Setting $\xi_i=f^*(X_i)-Y_i$, observe that by \eqref{eq:squared-loss},
\begin{equation*}
P_N \cL_f = \frac{1}{N} \sum_{i=1}^N (f-f^*)^2(X_i) + \frac{2}{N} \sum_{i=1}^N \xi_i (f-f^*)(X_i).
\end{equation*}
Hence, one may bound $P_N \cL_f$ from below by showing that with high probability, and in rather general situations,
\begin{description}
\item{1.} {\it The version space condition holds}: that is, if $\|f-f^*\|_{L_2} \geq \beta_N^*(\gamma_1)$ then
$$
\frac{1}{N} \sum_{i=1}^N (f-f^*)^2(X_i) \geq c_1\gamma_1 \|f-f^*\|_{L_2}^2.
$$
\item{2.} {\it The noise interaction condition holds}: that is, if $\|f-f^*\|_{L_2} \geq \alpha_N^*(\gamma_2,\delta)$, then
$$
\left|\frac{1}{N}\sum_{i=1}^N \xi_i (f-f^*)(X_i)\right| \leq c_2\gamma_2 \|f-f^*\|_{L_2}^2.
$$
\end{description}

Therefore, if $\gamma_2$ is chosen to be smaller than $c_1\gamma_1/2c_2$ and if $(X_i,Y_i)_{i=1}^N$ is a sample for which both (1) and (2) hold, then $P_N \cL_f >0$ when $\|f-f^*\|_{L_2} \geq \max\{\alpha_N^*(\gamma_2,\delta),\beta_N^*(\gamma_1)\}$. In particular, on that event, the estimation error satisfies
$$
\|\hat{f}-f^*\|_{L_2} \leq \max\{\alpha_N^*(\gamma_2,\delta),\beta_N^*(\gamma_1)\}.
$$

Let us emphasize that one may obtain a very good lower bound on the quadratic component of $P_N \cL_f$ (and thus on the version space condition), solely because the required estimate is {\it one-sided}. A two-sided bound, obtained by an upper estimate on centred quadratic process
$$
\sup_{f \in \cF \cap s\cD_{f^*}} \left|\frac{1}{N} \sum_{i=1}^N (f-f^*)^2(X_i) - \E (f-f^*)^2\right|,
$$
requires $\cF$ to be highly regular in some sense (see \cite{MPT,Men-weak,MenPao1,MenPao2} for two-sided estimates on the quadratic process).  And, as noted earlier, a two-sided bound of this type may be false even for a single function, let alone for a class of functions, unless one imposes rather restrictive assumptions.

\subsection{The assumption on the class} \label{sec:ass}
The key assumption leading the lower bound on the quadratic term is the following:
\begin{Assumption} \label{ass:small-ball}
Let $\cH \subset L_2(\mu)$ be a class of functions and set
$$
Q_{\cH} (u) = \inf_{h \in \cH} Pr(|h| \geq u \|h\|_{L_2} ).
$$
Given a class of functions $\cF$, let $\cF - \cF = \{f-h : f,h \in \cF\}$. We will assume that there is some $u>0$ for which $Q_{\cF-\cF}(u) > 0$.
\end{Assumption}
In Section \ref{sec:examples} we will present several generic examples showing that this weak {\it small-ball condition} is indeed minimal, and that in many cases one may choose $u$ and $Q$ to be appropriate absolute constants.

\vskip0.5cm
This notion of small-ball is very different from concentration. The empirical mean of a function will concentrate around its true mean only if the function has well-behaved high moments. Such a condition is trivially satisfied when the function is a bounded, but in general it is highly restrictive. In contrast, it is well known that any sort of moment equivalence, even as weak as $\|h\|_{L_2} \leq L \|h\|_{L_1}$, leads to the nontrivial small-ball estimate
\begin{equation} \label{eq:in-text-small-ball}
Pr(|h| \geq c_1(L)\|h\|_{L_2}) \geq c_2(L)
\end{equation}
for constants $c_1$ and $c_2$ that depend only on $L$ (see Section \ref{sec:examples} for more details).

The significance of \eqref{eq:in-text-small-ball} is that it may be used to derive a very high probability lower bound on $P_N h^2$. Indeed, let $X_1,...,X_N$ be independent, distributed according to $\mu$. A straightforward binomial estimate shows that with probability at least $1-2\exp(-c_3(L)N)$, at least $c_2(L)N/2$ of the $|h(X_i)|$'s are larger than $c_1(L)\|h\|_{L_2}$. Hence, on that event,
$$
\frac{1}{N}\sum_{i=1}^N h^2(X_i) \geq c_4(L) \E h^2.
$$

We are, at last, ready to formulate the main result of the article.
\begin{Theorem} \label{thm:main}
Let $\cF \subset L_2(\mu)$ be a closed, convex class of functions, set $Y$ to be the unknown target and put $\xi(X,Y)=f^*(X)-Y$.

Fix $\tau>0$ for which $Q_{\cF-\cF}(2\tau)>0$ and set $\gamma < \tau^2 Q_{\cF-\cF}(2\tau)/16$. For every $0<\delta<1$, with probability at least $1-\delta-\exp(-NQ_{\cF-\cF}^2(2\tau)/2)$ one has
$$
\|\hat{f}-f^*\|_{L_2} \leq 2\max\left\{\alpha_N^* \left(\gamma,\delta/4\right),\beta_N^* \left(\frac{\tau Q_{\cF-\cF}(2\tau)}{16}\right)\right\}.
$$
\end{Theorem}
The proof of Theorem \ref{thm:main} will be presented in Section \ref{sec:proof}.

\vskip0.5cm
Unlike Theorem \ref{thm:Kolt}, Theorem \ref{thm:main} holds with essentially no restrictions on the class or on the target. In particular, it may be applied in all the examples described in the introduction that fall outside the scope of Theorem \ref{thm:Kolt}, once $\cF-\cF$ satisfies Assumption \ref{ass:small-ball}.

\vskip0.5cm

To illustrate the clear advantages Theorem \ref{thm:main} has over Theorem \ref{thm:Kolt}, we will present one example in which $\alpha_N^*$ and $\beta_N^*$ may be computed in a relatively straightforward way: the persistence problem in $\ell_1^n$  \cite{GR,Gr,BMN}.

Consider a family of estimation problems in $\cF_R =\left\{\inr{t,\cdot} : t \in RB_1^n\right\}$ and for a set of reasonable targets. In the persistence framework, the dimension $n$, the radius $R$ and the noise level $\sigma$ are allowed to grow with the sample size $N$, and one has to find conditions on $n(N)$, $R(N)$ and $\sigma(N)$ that still ensure that $\|\hat{f}-f^*\|_{L_2}$ tends to zero with high probability. Therefore, one has to identify the correct way in which the estimation error scales with each one of these parameters.

We will study the persistence problem when $X$ has bounded iid coordinates and $Y=\inr{t_0,\cdot} +W$ for $t_0 \in RB_1^n$ and a bounded, symmetric random variable $W$ that is independent of $X$. Although the problem fits the assumptions of Theorem \ref{thm:Kolt}, the outcome of Theorem \ref{thm:main} turns out to be superior by far, and, in fact, leads to the optimal estimate in the minimax sense.

\subsection{Is Theorem \ref{thm:main} optimal?}
The question of whether Theorem \ref{thm:main} is optimal is rather delicate and requires highly technical machinery if it is to be answered in full. To keep the scope and length of the article within reason, we will only sketch a few facts indicating that Theorem \ref{thm:main} is optimal, at least for a wide variety of classes.

As noted earlier, the intrinsic parameter $\beta_N^*$, which characterizes the low-noise regime, is also an upper estimate on the diameter of the version space associated with $f^*$ in $L_2(\mu)$. There are many examples in which this estimate is sharp, including the class of linear functionals used in the persistence framework, endowed by $RB_1^n$.

As it turns out, the $L_2(\mu)$ diameter of the version space is a lower bound on the estimation error in a rather general sense and independently of the learning procedure that is used. Indeed, fix a mean-zero random variable $W$ that is independent of $X$, consider an arbitrary class $\cF$ and the family of targets $Y^f=f(X)+W$, where $f \in \cF$. Given $(X_i)_{i=1}^N$ and $f \in \cF$,
$$
{\cal V}\left(f,(X_i)_{i=1}^N\right)=\{h \in \cF : \ h(X_i)=f(X_i) \ {\rm for \ every \ } 1 \leq i \leq N\}
$$
is the version space associated with $(X_i)_{i=1}^N$ and $f$, and let ${\cal R}\left(f,(X_i)_{i=1}^N\right)$ be its diameter in $L_2(\mu)$. Let $\tilde{f}$ be any learning procedure that selects an element in $\cF$ based on the random data $(X_i,Y^f_i)_{i=1}^N$.

The following result shows that $\tilde{f}$ cannot perform well on all the learning problems defined by the targets $Y^f$:
\begin{Theorem} \label{thm:lower-diam-ver} \cite{LM1}
Let $\tilde{f}$ be a learning procedure as above. There exists some $f \in \cF$ for which
$$
Pr \left(\|\tilde{f}-f\|_{L_2} \geq \frac{1}{4}{\cal R}\left(f,(X_i)_{i=1}^N\right) \right) \geq \frac{1}{2},
$$
where the probability is taken with respect to the data $(X_i,Y_i^f)_{i=1}^N$.
\end{Theorem}

Therefore, $\beta_N^*$ captures the estimation error in the low-noise regime, with the exception of the (rare) examples in which it is far from the typical diameter of the version space ${\cal V}(f^*,(X_i)_{i=1}^N)$ in $L_2(\mu)$ .

\vskip0.5cm

As for $\alpha_N^*$, there are strong indications it is the right parameter for describing the interaction between the target and the class. Optimality questions of that flavour have been studied in \cite{LM1}, featuring arbitrary classes of function $\cF$ and targets $Y^f=f(X)+W$ for some $f \in \cF$ and mean-zero gaussian variables $W$ that are independent of $X$.

The results of \cite{LM1} show that a `subgaussian version' of $\alpha_N^*$ is optimal, under mild assumptions on the canonical gaussian process indexed by $\cF$. Without going into accurate and rather technical definitions, one may show that if $\cF$ and $\xi=f^*(X)-Y$ satisfy a subgaussian condition, and if the canonical gaussian process indexed by $\cF$, $\{G_f : f \in \cF\}$, is sufficiently regular, then
\begin{equation} \label{eq:in-proof-subgaussian-complexity}
s_N^*(\gamma)=\inf\left\{s>0 : \|\xi\|_{L_2} \cdot \E \sup_{f \in \cF \cap s\cD_{f^*}} G_f \leq \gamma s^2 \sqrt{N} \right\}
\end{equation}
is the optimal complexity parameter for the high-noise regime in such estimation problems, for a constant $\gamma$ that depends only on the subgaussian properties of $\cF$.

One may also show \cite{Men-52} that under the same regularity and subgaussian assumptions on the gaussian process $\{G_f : f \in \cF\}$, $\|\xi\|_{L_2} \E \sup_{f \in \cF \cap r\cD_{f^*}} G_f$ is equivalent with high probability and in expectation to
$$
\sup_{f \in \cF \cap r\cD_{f^*}} \frac{1}{\sqrt{N}} \sum_{i=1}^N \eps_i\xi_i (f-f^*)(X_i).
$$
Hence, for the right choices of $\delta$ and $\gamma^\prime$, $\alpha_N^*(\gamma^\prime,\delta)$ coincides with $s_N^*(\gamma)$, showing its optimality.

Unfortunately, extending this somewhat sketchy argument to the general case studied here, in which the subgaussian machinery is not at one's disposal, requires methods that are well beyond the scope of this article. It is therefore deferred to future work.

\section{Some Examples} \label{sec:examples}
 Let us turn to situations one is likely to encounter in a heavy-tailed framework: classes of functions for which one has almost no moment control, and therefore, its members do not exhibit any useful two-sided concentration of empirical means; nevertheless, one may obtain a small-ball estimate as in Assumption \ref{ass:small-ball}.
\begin{Lemma} \label{Lemma:small-ball-examples}
Let $\cF$ be a class of functions on a probability space $(\Omega,\mu)$.
\begin{description}
\item{1.} If  $\|f_1-f_2\|_{L_2} \leq \kappa_1 \|f_1-f_2\|_{L_1}$ for every $f_1,f_2 \in \cF$, then there are constants $c_1$ and $c_2$ that depend only on $\kappa_1$ for which $Q_{\cF-\cF}(c_1) \geq c_2$.
\item{2.} If there are $p>2$ and $\kappa_2$ for which $\|f_1-f_2\|_{L_p} \leq \kappa_2 \|f_1-f_2\|_{L_2}$ for every $f_1,f_2 \in \cF$, then there are constants $c_1$ and $c_2$ that depend only on $\kappa_2$ and $p$ for which $Q_{\cF-\cF}(c_1) \geq c_2$.
\end{description}
\end{Lemma}
Lemma \ref{Lemma:small-ball-examples} is an immediate outcome of the Paley-Zygmund inequality (see, e.g. \cite{dlPG}). For example, if $p>2$ and $\|h\|_{L_p} \leq \kappa_2 \|h\|_{L_2}$  then by the Paley-Zygmund inequality, for every $0<u<1$,
$$
Pr(|h| \geq u\|h\|_{L_2}) \geq \left(\frac{1-u^2}{\kappa_2^2}\right)^{\frac{p}{p-2}}.
$$
Thus, when applied to each $h=f_1-f_2$ and for $u=1/2$, it follows that $Q_{\cF-\cF}(1/2) \geq (3/4\kappa_2^2)^{p/(p-2)}$, and the small-ball condition holds uniformly in $\cF-\cF$ with reasonable constants.

\vskip0.5cm

This type of moment condition is particularly useful because it passes smoothly to product measures, in the following sense. Let $\zeta$ be a mean-zero, variance $1$ random variable and set $X=(\zeta_1,...,\zeta_n)$ to be a vector with independent coordinates, distributed according to $\zeta$. Clearly, $X$ is isotropic because $\E\inr{X,t}^2 = \E \sup_{i,j} \xi_i \xi_j t_i t_j = \|t\|_{\ell_2^n}^2$ for every $t \in \R^n$.
\begin{Lemma} \label{lemma:iid-coordinates}
Let $\zeta$ and $X$ be as above.
\begin{description}
\item{1.} Assume that there is some $\kappa_1>0$ for which $\|\zeta\|_{L_2} \leq \kappa_1 \|\zeta\|_{L_1}$. Then $\|\inr{t,X}\|_{L_2} \leq c_1\|\inr{t,X}\|_{L_1}$ for every $t \in \R^n$ and for a constant $c_1$ that depends only on $\kappa_1$.
\item{2.} If $\|\zeta\|_{L_p} \leq \kappa_2$ for some $p>2$ then   $\|\inr{t,X}\|_{L_p} \leq c_2\sqrt{p}\kappa_2\|\inr{t,X}\|_{L_2}$ for every $t \in \R^n$ and for an absolute constant $c_2$.
\end{description}
\end{Lemma}
The proof of Lemma \ref{lemma:iid-coordinates} is presented in Section \ref{sec:iid-coord}.

\vskip0.5cm

Lemma \ref{lemma:iid-coordinates} leads to many examples in which Theorem \ref{thm:main} may be applied. Indeed, let $X=(\zeta_i)_{i=1}^n$ be a random vector with independent coordinates distributed as $\zeta$, set $T \subset \R^n$ to be a closed, convex set, and put $\cF_T = \left\{ \inr{t,\cdot} : t \in T\right\}$. Consider a square-integrable target $Y$ and let $f^*=\inr{t^*,\cdot}$ be the unique minimizer in $\cF$ of the functional $f \to \E (f(X)-Y)^2$. Since $X$ is isotropic, it follows that for every $f_t = \inr{t,\cdot}$, $\|f_t-f^*\|_{L_2}=\|t-t^*\|_{\ell_2^n}$.

\begin{Corollary} \label{cor:moments}
If either one of the moment conditions of Lemma \ref{lemma:iid-coordinates} holds for $\zeta$, then with probability at least $1-\delta-\exp(-c_1N)$, ERM produced $\hat{t} \in T$ for which
$$
\|\hat{t}-t^*\|_{\ell_2^n} \leq 2\max\{\alpha_N^*(c_2,\delta/4),\beta_N^*(c_3)\}
$$
for appropriate constants $c_1$, $c_2$ and $c_3$ that depend only on $\kappa_1$ or on $\kappa_2$ and $p$.
\end{Corollary}
Needless to say that this problem falls outside the scope of Theorem \ref{thm:Kolt}, as functions in $\cF$ and $Y$ need not be bounded, nor do they necessarily have rapidly decaying tails.

\subsection{The Persistence framework}
Let us turn to an example that illustrates the striking difference between Theorem \ref{thm:Kolt} and Theorem \ref{thm:main}, even in a bounded scenario, and in which $\alpha_N^*$ and $\beta_N^*$ are not difficult to compute.

Let $X=(\zeta_i)_{i=1}^n$ be a random vector with independent coordinates distributed according to a mean-zero, variance $1$ random variable $\zeta$. To give Theorem \ref{thm:main} and Theorem \ref{thm:Kolt} a `level playing field', assume that $|\zeta| \leq \kappa$ almost surely.

For every $R \geq 1$ let $\cF_R = \left\{\inr{t,\cdot} : t \in R B_1^n\right\}$. For the sake of simplicity, assume further that the unknown target is $Y=\inr{t_0,\cdot}+W$, for some $t_0 \in RB_1^n$ and a mean-zero random variable $W$ that has variance $\sigma \leq R$ and is independent of $X$. We will identify $\cF_R$ with $R B_1^n = \{ t \in \R^n : \|t\|_{\ell_1^n} \leq R\}$ in the natural way.

\begin{Problem} \label{Qu-persistence}
If $\hat{t} \in RB_1^n$ is selected by ERM using an $N$-sample $(X_i,Y_i)_{i=1}^N$, find a function $\rho(N,n,R,\sigma,\delta)$ for which $\|\hat{t}-t_0\|_{\ell_2^n} \leq \rho$ with probability at least $1-\delta$.
\end{Problem}

Recall that Problem \ref{Qu-persistence} has been mentioned in the introduction for $X=(\eps_1,...,\eps_N)$ and $W=\sigma \eps_{N+1}$.

The following two statements summarize the outcomes of Theorem \ref{thm:Kolt} and Theorem \ref{thm:main} for this choice of $\cF$ and $Y$. The proofs of the claims may be found in Section \ref{app:per}.

Let us begin with the outcome of Theorem \ref{thm:Kolt}:
\begin{Theorem} \label{thm:persist-from-Kolt}
For every $\kappa>1$ there exist constants $c_1,c_2$ and $c_3$ that depend only on $\kappa$ for which the following holds. Assume that $\|\zeta\|_{L_\infty}, \|W\|_{L_\infty} \leq \kappa$. Set
$$
\rho_N=
\begin{cases}
\frac{R^2}{\sqrt{N}} \sqrt{\log \left(\frac{2c_1 n}{\sqrt{N}}\right)} & \mbox{if} \ \ N \leq c_1 n^2
\\
\\
\frac{R^2n}{N} & \mbox{if} \ \ N >c_1 n^2.
\end{cases}
$$
Then, with probability at least $1-2\exp(- c_2N \rho_N/R^2)$, ERM produces $\hat{t}$ that satisfies $\|\hat{t}-t_0\|_{\ell_2^n}^2 \leq c_3\rho_N$.
\end{Theorem}
The estimation error in Theorem \ref{thm:persist-from-Kolt} does not scale correctly with $R$ ($R^2/\sqrt{N}$ is clearly too big) and also does not depend on any $L_p$ norm of the noise $W$, except for the trivial bound $\|W\|_{L_\infty}=\kappa$, which is likely to be much larger than, say, the variance $\|W\|_{L_2}$.

In contrast, the next result follows from Theorem \ref{thm:main}. To formulate it, set
$$
\|W\|_{L_{2,1}} = \int_0^\infty \sqrt{Pr (|W| >t)} dt.
$$
It turns out that $\|W\|_{L_{2,1}}$, which is slightly larger than $\|W\|_{L_2}$ but smaller than $c(q)\|W\|_{L_q}$ for any $q>2$, captures the noise level of the problem. Since in virtually all examples the $L_2$ and $L_{2,1}$ norms are equivalent, we will abuse notation and denote $\sigma=\|W\|_{L_{2,1}}$ rather than $\sigma=\|W\|_{L_2}$.

\begin{Theorem} \label{thm:persist-from-main}
For every $\kappa>1$ there exist constants $c_1,c_2,c_3$ and $c_4$ that depend only on $\kappa$ for which the following holds. Assume that $\|\zeta\|_{L_\infty}, \|W\|_{L_\infty} \leq \kappa$ and that $\|W\|_{2,1} = \sigma <\infty$. Put
\begin{equation*}
v_1=
\begin{cases}
\frac{R^2}{N} \log\left(\frac{2c_1n}{N}\right) &  \mbox{if} \ \ N \leq c_1 n,
\\
\\
0 & \mbox{if} \ \ N > c_2 n.
\end{cases}
\end{equation*}
and
\begin{equation*}
v_2 =
\begin{cases}
\frac{R\sigma}{\sqrt{N}} \sqrt{\log\left(\frac{2c_2n\sigma}{\sqrt{N}R}\right)} & \mbox{if } \ \  N \leq c_2n^2 \sigma^2/R^2
\\
\\
\frac{\sigma^2 n}{N} & \mbox{if} \ \ N >c_2n^2 \sigma^2/R^2.
\end{cases}
\end{equation*}
Then with probability at least
$$
1-2\exp\left(-c_3N v_2 \min\left\{\frac{1}{\sigma^2},\frac{1}{R}\right\}\right),
$$
$$
\|\hat{t}-t_0\|_{\ell_2^n}^2 \leq c_4\max\left\{ v_1,v_2\right\}.
$$
\end{Theorem}
Theorem \ref{thm:persist-from-main} yields a much better dependence of $\|\hat{t}-t_0\|_{\ell_2^n}^2$ on the parameters involved than Theorem \ref{thm:persist-from-Kolt}. We will show that $v_1$ is an upper bound on $\beta_N^*$, and thus on the diameter of the version space, while $v_2$ is an upper bound on $\alpha_N^*$, capturing the interaction class members have with the noise.

The results of \cite{LM1} show that up to the exact probability estimate, the estimation error given in Theorem \ref{thm:persist-from-main} is optimal in the minimax sense, and no procedure can perform with confidence larger than $3/4$ and yield a estimation error better than $\sim \max\{v_1,v_2\}$.

It turns out that Theorem \ref{thm:persist-from-main} can be extended even further. For example, it holds for a general target $Y$ rather than just for $Y=\inr{t_0,\cdot}+W$, the assumption that $X$ has iid coordinates can be relaxed, and `heavy-tailed' measures may be used instead. Unfortunately, the proof of a more general version of Theorem \ref{thm:persist-from-main} comes at a high technical cost and we refer the reader to \cite{LM-lasso} and \cite{Men-extend-ver} for more details.

\section{Proof of Theorem \ref{thm:main}} \label{sec:proof}
We begin this section with a few definitions that will be needed in the proof of Theorem \ref{thm:main}.

\begin{Definition} \label{def:star-shaped}
A class $\cH$ is star-shaped around $0$ if for every $h \in \cH$ and any $0<\lambda \leq 1$, $\lambda h \in \cH$. Thus, a class is star-shaped around $0$ if it contains the entire interval $[0,h]$ whenever $h \in \cH$.
\end{Definition}

\begin{Definition} \label{def:fixed-point}
For every $\gamma>0$, set
$$
\beta_N(\cH,\gamma) = \inf\left\{r>0: \E \sup_{h \in \cH \cap r\cD} \left|\frac{1}{N}\sum_{i=1}^N \eps_i h(X_i)\right| \leq \gamma r \right\}.
$$

We will sometimes write $\beta_N(\gamma)$ instead of $\beta_N(\cH,\gamma)$.
\end{Definition}

Let $\cF-f^*=\{f-f^* : f \in \cF\}$ and observe that $\beta_N^*(\gamma)=\beta_N(\cF-f^*,\gamma)$. Also, it is straightforward to verify that if $\cH$ is star-shaped around $0$ and $r>\beta_N(\gamma)$ then
$$
\E \sup_{h \in \cH \cap r\cD} \left| \frac{1}{N} \sum_{i=1}^N \eps_i h(X_i) \right| \leq \gamma r
$$
(see, for example, the discussion in \cite{LM1}).

The main component in the proof of Theorem \ref{thm:main} is the following:
\begin{Theorem} \label{thm:quad-est}
Let $\cF \subset L_2(\mu)$ be a closed, convex class and assume that there is some $\tau>0$ for which $Q_{\cF-\cF}(2\tau)>0$. Given $f^* \in \cF$, set $\cH=\cF-f^*$. Then, for every $r>\beta_N(\cH,\tau Q_{\cH}(2\tau)/16)$, with probability at least $1-2\exp(-NQ_{\cH}^2(2\tau)/2)$, if $\|f-f^*\|_{L_2} \geq r$, one has
\begin{equation} \label{eq:large-subset}
\left|\left\{i : |(f-f^*)(X_i)| \geq \tau \|f-f^*\|_{L_2} \right\} \right| \geq N \frac{Q_{\cH}(2\tau)}{4}.
\end{equation}
\end{Theorem}

\vskip0.5cm

The first step in the proof of Theorem \ref{thm:quad-est} is the following uniform empirical small-ball estimate -- which is of a similar nature to the results from \cite{KM} and \cite{Men-section}.

\begin{Theorem} \label{thm-small-ball-vlad}
Let $S(L_2)$ be the $L_2(\mu)$ unit sphere and let ${\cal H} \subset S(L_2)$. Assume that there is some $\tau>0$ for which $Q_{\cH}(2\tau)>0$. If
$$
\E \sup_{h \in {\cal H}} \left|\frac{1}{N}\sum_{i=1}^N \eps_i h(X_i) \right| \leq \frac{\tau Q_{\cH}(2\tau)}{16},
$$
then with probability at least $1-2\exp(-NQ_{\cH}^2(2\tau)/2)$,
$$
\inf_{h \in {\cal H}} |\{i : |h(X_i)| \geq \tau\}| \geq  N \frac{Q_{\cH}(2\tau)}{4}.
$$
\end{Theorem}

\proof
Recall that $P_Nf = \frac{1}{N} \sum_{i=1}^N f(X_i)$ and $P f = \E f(X)$, and note that for every $h \in {\cal H}$ and $u>0$, $|\{i : |h(X_i)| \geq u\}| = NP_N \IND_{\{|h| \geq u\}}$. Also,
\begin{equation*}
P_N \IND_{\{|h| \geq u\}} = P \IND_{\{|h| \geq 2u\}} + \left(P_N \IND_{\{|h| \geq u\}}-P \IND_{\{|h| \geq 2u\}}\right)=(*).
\end{equation*}
Let $\phi_u:\R_+ \to [0,1]$ be the function
\begin{equation*}
\phi_u(t) =
\begin{cases}
1 &  \ \ t \geq 2u,
\\
(t/u)-1 & \ \ u \leq t \leq 2u,
\\
0 & \ \ t<u,
\end{cases}
\end{equation*}
and observe that for every $t \in \R$, $\IND_{[u,\infty)}(t) \geq \phi_u(t)$ and $\phi_u(t) \geq \IND_{[2u,\infty)}(t)$. Hence,
\begin{align*}
(*) & \geq P \IND_{\{|h| \geq 2u\}} + P_N \phi_u(|h|) - P \phi_u(|h|)
\\
& \geq  \inf_{h \in {\cal H}} Pr(|h| \geq 2u) - \sup_{h \in {\cal H}} \left|P_N \phi_u(|h|) - P \phi_u(|h|)\right|.
\end{align*}
Let $Z(X_1,...,X_N) = \sup_{h \in {\cal H}} \left|P_N \phi_u(|h|) - P \phi_u(|h|)\right|$. By the bounded differences inequality applied to $Z$
(see, for example, \cite{BLM}), it follows that for every $t>0$, with probability at least $1-2\exp(-2t^2)$,
$$
\sup_{h \in {\cal H}} \left|P_N \phi_u(|h|) - P \phi_u(|h|)\right| \leq \E \sup_{h \in {\cal H}} \left|P_N \phi_u(|h|) - P \phi_u(|h|)\right| + \frac{t}{\sqrt{N}}.
$$
Note that $\phi_u$ is a Lipschitz function that vanishes in $0$ and with a Lipschitz constant $1/u$. Therefore, by the Gin\'{e}-Zinn symmetrization theorem \cite{GZ84} and the contraction inequality for Bernoulli processes (see, e.g. \cite{LT}),
$$
\E \sup_{h \in {\cal H}} \left|P_N \phi_u(|h|) - P \phi_u(|h|)\right| \leq \frac{4}{u} \E \sup_{h \in {\cal H}} \left|\frac{1}{N} \sum_{i=1}^N \eps_i h(X_i) \right|.
$$
Hence, with probability at least $1-2\exp(-2t^2)$, for every $h \in {\cal H}$,
$$
P_N \IND_{\{|h| \geq u\}} \geq  \inf_{h \in {\cal H}} Pr(|h| \geq 2u) - \frac{4}{u} \E \sup_{h \in {\cal H}} \left|\frac{1}{N} \sum_{i=1}^N \eps_i h(X_i) \right| - \frac{t}{\sqrt{N}}.
$$
If $Q_{\cH}(2\tau)>0$, set $u=\tau$ and $t=\sqrt{N}Q_{\cH}(2\tau)/2$. Since
$$
\E \sup_{h \in {\cal H}} \left|\frac{1}{N} \sum_{i=1}^N \eps_i h(X_i) \right| \leq \frac{\tau Q_{\cH}(2\tau)}{16},
$$
it follows that with probability at least $1-2\exp(-NQ_{\cH}^2(2\tau)/2)$,
$$
|\{i : |h(X_i)| \geq \tau\}| \geq N \frac{Q_{\cH}(2\tau)}{4}.
$$
\endproof

\begin{Corollary} \label{cor:coordintes-star-shaped}
Let $\cH$ be star-shaped around $0$ and assume that there is some $\tau>0$ for which $Q_{\cH}(2\tau)>0$. Then, for every $r>\beta_N(\cH, \tau Q_{\cH}(2\tau)/16)$,
with probability at least $1-2\exp(-NQ_{\cH}^2(2\tau)/2)$, for every $h \in \cH$ that satisfies $\|h\|_{L_2} \geq r$,
\begin{equation} \label{eq:in-cor-star}
|\{i : |h(X_i)| \geq \tau\|h\|_{L_2}\}| \geq  N \frac{Q_{\cH}(2\tau)}{4}.
\end{equation}
\end{Corollary}

\proof
Let $r>\beta_N(\cH, \tau Q_{\cH}(2\tau)/16)$ and since $\cH$ is star-shaped around $0$,
$$
\E \sup_{h \in \cH \cap r \cD} \left|\frac{1}{N}\sum_{i=1}^N \eps_i h(X_i) \right| \leq \frac{\tau Q_{\cH}(2\tau)}{16} r.
$$
Consider the set
$$
V=\left\{h/r : h \in \cH \cap r S(L_2)\right\} \subset S(L_2).
$$
Clearly, $Q_V(2\tau) \geq Q_{\cH}(2\tau)$ and
$$
\E \sup_{v \in V} \left| \frac{1}{N} \sum_{i=1}^N \eps_i v(X_i) \right| =\E \sup_{h \in \cH \cap r S(L_2)} \left|\frac{1}{N}\sum_{i=1}^N \eps_i \frac{h(X_i)}{r}\right| \leq \frac{\tau Q_{\cH}(2\tau)}{16} \leq \frac{\tau Q_{V}(2\tau)}{16}.
$$
By Theorem \ref{thm-small-ball-vlad} applied to the set $V$ and since $Q_V(2\tau) \geq Q_{\cH}(2\tau)$, it follows that with probability at least $1-2\exp(-NQ_{\cH}^2(2\tau)/2)$, for every $v \in V$,
$$
|\{i : |v(X_i)| \geq \tau\}| \geq  N \frac{Q_{\cH}(2\tau)}{4}.
$$
Next, fix any $h \in \cH$ for which $\|h\|_{L_2} \geq r$. $\cH$ is star-shaped around $0$ and thus $(r/\|h\|_{L_2}) h \in \cH \cap r S(L_2)$, implying that $h/\|h\|_{L_2} \in V$. The claim follows because \eqref{eq:in-cor-star} is positive homogeneous,
\endproof

\noindent {\bf Proof of Theorem \ref{thm:quad-est}.} Given $f^* \in \cF$, let $\cH = \cF - f^*=\{f-f^* : f \in \cF\}$. Applying Corollary \ref{cor:coordintes-star-shaped}, it suffices to show that $\cH$ is star-shaped around $0$. To that end, observe that if $f-f^* \in \cH$ and $0  \leq \lambda \leq 1$, then $\lambda (f-f^*) = w-f^*$ for $w=\lambda f + (1-\lambda)f^*$. Hence, as $\cF$ is convex, $w \in \cF$.
\endproof

\noindent{\bf Proof of Theorem \ref{thm:main}.} Recall that $\xi(X,Y)=f^*(X)-Y$, that $(\xi_i)_{i=1}^N=(f^*(X_i)-Y_i)_{i=1}^N$, and that for every $f \in \cF$,
$$
P_N \cL_f = \frac{2}{N}\sum_{i=1}^N \xi_i (f-f^*)(X_i) + \frac{1}{N} \sum_{i=1}^N (f-f^*)^2 (X_i).
$$
The function $f^*$ minimizes in $\cF$ the distances $\|Y-f(X)\|_{L_2}$, and by the characterization of the metric projection onto a convex set in a Hilbert space, $\E \xi(f-f^*)(X) \geq 0$.

Since $\cH \subset \cF-\cF$, the assertion of Theorem \ref{thm:quad-est} for $\cH = \cF-f^*$ holds with $Q_{\cF-\cF}(2\tau)$ replacing the larger $Q_{\cH}(2\tau)$. Hence, if $r>\beta_N(\cH,\tau Q_{\cF-\cF}(2\tau)/16)$, then with probability at least $1-2\exp(-NQ_{\cF-\cF}^2(2\tau)/2)$, for every $f \in \cF$ that satisfies $\|f-f^*\|_{L_2} \geq  r$, one has
\begin{equation} \label{eq:quadratic}
\frac{1}{N} \sum_{i=1}^N (f-f^*)^2(X_i) \geq \frac{\tau^2}{4} Q_{\cF-\cF}(2\tau) \cdot \|f-f^*\|_{L_2}^2.
\end{equation}
Therefore, on that event, if $\|f-f^*\|_{L_2} \geq r$,
\begin{equation*}
P_N \cL_f \geq \left(\frac{1}{N} \sum_{i=1}^N \xi_i(f-f^*)(X_i) - \E \xi(f-f^*) \right) + \frac{\tau^2}{4}Q_{\cF-\cF}(2\tau) \cdot \|f-f^*\|_{L_2}^2.
\end{equation*}

Fix $\gamma$ to be named later and consider $\alpha_N^*(\gamma,\delta/4) \equiv \alpha_N$. Thus, with probability at least $1-\delta/4$, if $\|f-f^*\|_{L_2} \leq \alpha_N$ then
$$
\left|\frac{1}{N} \sum_{i=1}^N \eps_i \xi_i (f-f^*)(X_i) \right| \leq \gamma \alpha_N^2,
$$
and by the Gin\'{e}-Zinn symmetrization theorem \cite{GZ84}, with probability at least $1-\delta$,
\begin{equation} \label{eq:in-proof-linear-process}
\sup_{f-f^* \in (\cF-f^*) \cap \alpha_N S(L_2)} \left|\frac{1}{N} \sum_{i=1}^N \xi_i(f-f^*)(X_i)-\E\xi (f-f^*) \right| \leq 4\gamma \alpha_N^2.
\end{equation}
 Fix a sample for which \eqref{eq:in-proof-linear-process} holds and consider $f \in \cF$ that satisfies $\|f-f^*\|_{L_2} \geq \alpha_N$. Since $\cF-f^*$ is star-shaped around $0$ and  
 $$
 (\alpha_N/\|f-f^*\|_{L_2}) \cdot (f-f^*) \in (\cF-f^*) \cap \alpha_N S(L_2),
 $$ 
 one has
\begin{equation} \label{eq:linear}
\left|\frac{1}{N} \sum_{i=1}^N \xi_i(f-f^*)(X_i)-\E\xi(f-f^*) \right| \leq 4\gamma \alpha_N^2 \cdot \frac{ \|f-f^*\|_{L_2}}{\alpha_N} \leq 4\gamma \|f-f^*\|_{L_2}^2.
\end{equation}
Combining \eqref{eq:linear} and \eqref{eq:quadratic}, it is evident that with probability at least
$$
1-\delta-2\exp(-NQ_{\cF-\cF}^2(2\tau)/2),
$$
if $\|f-f^*\|_{L_2} \geq \max\{\alpha_N, r\}$ then
$$
P_N \cL_f \geq \|f-f^*\|_{L_2}^2 \left(-4\gamma +\frac{\tau^2Q_{\cF-\cF}(2\tau)}{4}\right)>0,
$$
provided that $\gamma < \tau^2 Q_{\cF-\cF}(2\tau)/16$. On that event,
$$
\|\hat{f}-f^*\|_{L_2} \leq 2\max\left\{ \alpha_N^*\left(\gamma,\delta/4\right), r \right\},
$$
as claimed.
\endproof

\section{Additional proofs} \label{sec:add-proof}
Here, we will present proofs of some of the claims that have been formulated in previous sections.

\subsection{Proof of Lemma \ref{lemma:iid-coordinates}} \label{sec:iid-coord}
We will prove a stronger statement: that the equivalence of the $p$-th moment and the second moment of the linear forms $\inr{X,t}$ holds for any isotropic random vector $X$ that is also unconditional and whose coordinates belong to $L_p$ for a fixed $p>2$. That is, when $X=(x_1,...,x_n)$ has the same distribution as $(\eps_1 x_1,...,\eps_n x_n)$ for every choice of signs $(\eps_i)_{i=1}^n$ (unconditionality), $\|x_i\|_{L_2}=1$ and $\|x_i\|_{L_p} \leq \kappa$.

Such a random vector can be heavy-tailed, as its coordinates may only belong to an $L_p$ space for a fixed $p>2$ that is close to $2$. Thus, the empirical mean of, say, $\inr{X,e_j}^2=x_j^2$ may exhibit rather poor concentration around the true mean, which is $1$. Despite that, the claim is that $\|\inr{X,t}\|_{L_p} \leq c \sqrt{p} \kappa \|\inr{X,t}\|_{L_2}$, and any class of linear functionals satisfies a small-ball condition with constants that depend only on $\kappa$ and $p$.

Clearly, a possible choice of $X$ is a random vector with independent, symmetric, variance one coordinates, and a standard symmetrization argument may be used to show that the same holds when the coordinates are mean-zero rather than symmetric. Thus, the unconditional case extends Lemma \ref{lemma:iid-coordinates}.
\begin{Lemma} \label{lemma:unconditional}
There is an absolute constant $c$ for which the following holds. Let $X=(x_i)_{i=1}^n$ be an isotropic, unconditional vector in $\R^n$ and assume that $\max_{1 \leq i \leq n} \|x_i\|_{L_p} \leq \kappa$ for some $p>2$. Then, for every $t \in \R^n$,
$$
\|\inr{X,t}\|_{L_p} \leq c \sqrt{p} \kappa \|\inr{X,t}\|_{L_2}.
$$
\end{Lemma}

\proof Fix $t \in \R^n$. By the unconditionality of $X$ followed by Khintchine's inequality, there is an absolute constant $c$ for which
\begin{equation*}
\E|\sum_{i=1}^n t_i x_i|^p = \E_X \left(\E_\eps |\sum_{i=1}^n \eps_i t_i x_i |^p \right) \leq c^p p^{p/2} \E \left(\sum_{i=1}^n t_i^2 x_i^2 \right)^{p/2}.
\end{equation*}
Observe that for every $q \geq 1$ and every $y_1,...,y_n \in L_q$, the function $u \to \|\sum_{i=1}^n u_i y_i\|_{L_q}$ is a convex function in $u$, and thus it attains its maximum in the unit ball of $\ell_1^n$ in an extreme point. Therefore, setting $y_i=x_i^2$ and $q=p/2>1$,
$$
\sup_{t \in S^{n-1}} \E \left(\sum_{i=1}^n t_i^2 x_i^2 \right)^{p/2} =\left(\sup_{u \in B_1^n} \|\sum_{i=1}^n u_i y_i \|_{L_{p/2}} \right)^{p/2}
\leq \max_{1 \leq i \leq n} \E|x_i|^p \leq \kappa^p.
$$
Since $X$ is isotropic, $\|\inr{X,t}\|_{L_2} = \|t\|_{\ell_2^n}$, implying that
$$
\sup_{t \in S^{n-1}} \|\inr{X,t}\|_{L_p} \leq c \sqrt{p} \kappa=c \sqrt{p} \kappa \|\inr{X,t}\|_{L_2}.
$$
\endproof

\begin{Lemma} \label{lemma:second-equivalence}
Let $X=(x_i)_{i=1}^n$ be an isotropic, unconditional vector and assume that for every $1 \leq i \leq n$, $Pr(|x_i| \geq \lambda) \geq \kappa$. Then, for every $t \in \R^n$,
$$
\|\inr{X,t}\|_{L_2} \leq c(\lambda,\kappa) \|\inr{X,t}\|_{L_1},
$$
and $c(\lambda,\kappa)$ is a constant that depends only on $\lambda$ and $\kappa$.
\end{Lemma}
\proof Since $X$ is isotropic, $\|\inr{X,t}\|_{L_2}=\|t\|_{\ell_2^n}$, and since it is unconditional, by Khintchine's inequality, there is an absolute constant $c_1$ for which
\begin{align*}
\|\inr{X,t}\|_{L_1} = & \E_X \E_\eps \left|\sum_{i=1}^n \eps_i x_i t_i \right| \geq c_1 \E_X \left(\sum_{i=1}^n x_i^2 t_i^2 \right)^{1/2}
\\
\geq & c_1 \E_X \left(\sum_{i=1}^n x_i^2 \IND_{\{|x_i| \geq \lambda\}}  t_i^2 \right)^{1/2} \geq c_1 \lambda \E \left(\sum_{i=1}^n \delta_i t_i^2 \right)^{1/2},
\end{align*}
where $\delta_i$ are $\{0,1\}$-valued random variables that satisfy $\E \delta_i = Pr(|x_i| \geq \lambda) \geq \kappa$.

Set $Z=\sum_{i=1}^n \delta_i t_i^2$. Observe that $\E Z \geq \kappa \|t\|_{\ell_2^n}^2$ and since $\delta_i \leq 1$, $\E Z^2 \leq \|t\|_{\ell_2^n}^4$. Thus $\|Z\|_{L_1}/\|Z\|_{L_2} \geq \kappa$ and by a standard application of the Paley-Zygmund inequality,
$Pr\left(Z \geq \|Z\|_{L_1}/{2}\right) \geq \kappa^2/4$. Therefore,
$$
\E \left(\sum_{i=1}^n \delta_i t_i^2 \right)^{1/2} \geq \frac{\kappa^{5/2}}{4 \sqrt{2}} \|t\|_{\ell_2^n}=\frac{\kappa^{5/2}}{4 \sqrt{2}} \|\inr{X,t}\|_{L_2},
$$
implying that $\|\inr{X,t}\|_{L_1} \geq c(\lambda,\kappa) \|\inr{X,t}\|_{L_2}$.
\endproof

\subsection{The persistence problem} \label{app:per}
The proofs of Theorem \ref{thm:persist-from-Kolt} and Theorem \ref{thm:persist-from-main} follow from several observations.

\begin{Definition} \label{def:psi-2}
The $\psi_2$ norm of a random variable $X$ is
$$
\|X\|_{\psi_2} = \inf \left\{ c>0 : \E \exp(|X|^2/c^2) \leq 2 \right\}.
$$
A random variable $X$ is called $L$-subgaussian if $\|X\|_{\psi_2} \leq L \|X\|_{L_2}$.
\end{Definition}
It is standard to verify that $\|X\|_{\psi_2}$ is equivalent to the smallest constant $\kappa$ for which $Pr(|X| \geq t ) \leq 2\exp(-t^2/2\kappa^2)$ for every $t \geq 1$. Also, $\|X\|_{\psi_2}$ is equivalent to $\sup_{p \geq 2} \|X\|_{L_p}/\sqrt{p}$ (see, e.g. \cite{LAMA}).

Another property of $\psi_2$ random variables is that if $X_1,...,X_N$ are independent and mean-zero, then for every $a_1,...,a_N$,
$$
\|\sum_{i=1}^N a_i X_i\|_{\psi_2} \leq C \left(\sum_{i=1}^N a_i^2 \|X_i\|_{\psi_2}^2 \right)^{1/2}
$$
for a suitable absolute constant $C$.

\vskip0.5cm

our first observation is well known and follows because the intersection of the Euclidean unit ball $B_2^n$ and the $\ell_1^n$ ball of radius $\sqrt{d}$ is equivalent to the convex hull of vectors on the sphere, supported on at most $d$ coordinates (see, e.g., \cite{MPT}).

\begin{Lemma} \label{lemma:localized-short-supp}
If $Z=(z_i)_{i=1}^n$ is a random vector on $\R^n$, then for every integer $1 \leq d \leq n$,
$$
\E \sup_{t \in \sqrt{d}B_1^n \cap B_2^n} \inr{Z,t} \leq 2\E \left(\sum_{i=1}^{d} (z_i^*)^2 \right)^{1/2},
$$
where $(z_i^*)_{i=1}^n$ is a monotone non-increasing rearrangement of $(|z_i|)_{i=1}^n$.
\end{Lemma}

In view of Lemma \ref{lemma:localized-short-supp}, let us estimate
$\E \left(\sum_{i=1}^{d} (z_i^*)^2 \right)^{1/2}$.
\begin{Lemma} \label{lemma:d-sum}
There exists an absolute constant $C$ for which the following holds.
Assume that $z_1,...,z_n$ are independent copies of a mean-zero, variance $1$ random variable $z$, and that for every $p \leq \log n$, $\|z\|_{L_p} \leq \kappa \sqrt{p}$. Then for every $1 \leq d \leq n$,
$$
\E\left(\sum_{i=1}^d (z_i^*)^2 \right)^{1/2} \leq C\kappa \sqrt{d \log(en/d)}.
$$
\end{Lemma}

\proof For every $1 \leq j \leq n$ and $p \geq 2$,
$$
Pr(z_j^* \geq t) \leq \binom{n}{j}Pr^j(|z|>t) \leq \binom{n}{j}\left(\frac{\|z\|_{L_p}}{t}\right)^{jp}.
$$
Therefore, if $t=u\kappa\sqrt{\log(en/j)}$ and $p=\log(en/j)$ then
$$
Pr\left(z_j^* \geq u \kappa \sqrt{\log(en/j)}\right) \leq (e/u)^{j\log(en/j)}.
$$
Integrating the tails, $\E (z_j^*)^2 \leq c_1 \kappa^2 \log(en/j)$, and an application of Jensen's inequality completes the proof.
\endproof

The final component needed in the proofs of the two theorems is the following. Note that if $\cF_R = \{\inr{t,\cdot} : t \in R B_1^n\}$ and $Y=\inr{t_0,\cdot} + W$ for $t_0 \in RB_1^n$ and $W$ that is independent of $X$, then $f^*=\inr{t_0,\cdot}$. Also, by the convexity and symmetry of $B_1^n$,
$$
\left\{f-f^* : f \in \cF_R \cap s \cD_{f^*}\right\} \subset \left\{ \inr{t, \cdot} : t \in 2RB_1^n \cap s B_2^n \right\}.
$$
Therefore, if $X$ is a random vector with independent, mean-zero, variance 1 coordinates $\zeta_i$, and $X_1,...,X_N$ are independent copies of $X$, then
\begin{align} \label{eq:app-loc-avg-RB1-sB2}
& \E \sup_{f \in \cF_R \cap s \cD_{f^*}} \left|\frac{1}{\sqrt{N}}\sum_{i=1}^N \eps_i (f-f^*)(X_i) \right| \leq \E \sup_{t \in 2RB_1^n \cap s B_2^n} \inr{\frac{1}{\sqrt{N}}\sum_{i=1}^N X_i,t} \nonumber
\\
= & \E \sup_{t \in 2RB_1^n \cap s B_2^n} \inr{Z,t},
\end{align}
where $Z=(z_i)_{i=1}^n$ has independent coordinates. Each $z_i$ is distributed as $N^{-1/2}\sum_{j=1}^N \zeta_{i,j}$, and $(\zeta_{i,j})_{j=1}^N$ are independent copies of $\zeta_i$. Hence, by Lemma \ref{lemma:localized-short-supp},
$$
\E \sup_{t \in 2RB_1^n \cap s B_2^n} \inr{Z,t} = s \E\sup_{t \in (2R/s)B_1^n \cap  B_2^n} \inr{Z,t} \leq 2s \E \left(\sum_{i=1}^{(2R/s)^2} (z_i^*)^2 \right)^{1/2}.
$$
If the $\zeta_i$'s are distributed according to a mean-zero and variance $1$ random variable $\zeta$ that is bounded in $L_\infty$ by $\kappa$, then $\|\zeta\|_{\psi_2} \leq \kappa\|\zeta\|_{L_2}$ and $\zeta$ is $\kappa$-subgaussian. Moreover, $\|z_i\|_{L_2}=1$, and as a weighted sum of independent $\psi_2$ random variables, $\|z_i\|_{\psi_2} \leq c \kappa $ for a suitable absolute constant $c$. Hence, for every $p \geq 2$, $\|z_i\|_{L_p} \leq c_1 \kappa \sqrt{p}$.

Applying Lemma \ref{lemma:d-sum},
$$
s \E \left(\sum_{i=1}^{(2R/s)^2} (z_i^*)^2 \right)^{1/2} \leq c_2 \kappa s \cdot \frac{R}{s}\sqrt{\log (ens^2/R^2)} = c_2 \kappa R \sqrt{\log (ens^2/R^2)},
$$
provided that $(R/s)^2 \leq n/4$. Therefore, in that range,
\begin{equation} \label{eq:app-small-N}
\E \sup_{f \in \cF_R \cap s \cD_{f^*}} \left|\frac{1}{\sqrt{N}}\sum_{i=1}^N \eps_i (f-f^*)(X_i) \right|  \leq c_3 \kappa R \sqrt{\log (e\sqrt{n}s/R)},
\end{equation}
while if $(R/s)^2 \geq n/4$ one may verify that
\begin{equation} \label{eq:app-large-N}
\E \sup_{f \in \cF_R \cap s \cD_{f^*}} \left|\frac{1}{\sqrt{N}}\sum_{i=1}^N \eps_i (f-f^*)(X_i) \right| \leq c_3 \kappa s \sqrt{n}
\end{equation}
for a suitable absolute constant $c_3$.

\noindent{\bf Proof of Theorem \ref{thm:persist-from-Kolt}.} Note that $\|\zeta\|_{L_\infty} \leq \kappa$ and thus $\|\inr{t,X}\|_{L_\infty} \leq \kappa \|t\|_{\ell_1^n}  \leq \kappa R$. Hence, to make the class $\cF_R$ fit the bounded scenario, one has to scale down class members by a factor of $\kappa R $ and to consider the class
$$
\cF = \left\{\inr{t/\kappa R,\cdot} : t \in R B_1^n\right\}
$$
and the scaled-down target $\inr{t_0/\kappa R, \cdot} + W/\kappa R$.
If, by applying Theorem \ref{thm:persist-from-Kolt} to this scaled-down problem, one can show that $\|\hat{f}-f^*\|_{L_2}^2 \leq \rho$, then on the same event, $\|\hat{t}-t_0\|_{\ell_2^n}^2 \leq \kappa^2 R^2 \rho$.

Combining \eqref{eq:app-small-N} with Theorem \ref{thm:Kolt}, there is a high probability event on which, if $N \leq n^2$,
$$
\|\hat{t}-t_0\|_{\ell_2^n}^2 \leq c(\kappa)\frac{ R^2}{\sqrt{N}} \sqrt{\log(en/\sqrt{N})}
$$
and if $N>n^2$,
$$
\|\hat{t}-t^*\|_{\ell_2^n}^2 \leq c(\kappa) R^2 \frac{n}{N},
$$
as claimed.
\endproof
\vskip0.3cm
\noindent{\bf Proof of Theorem \ref{thm:persist-from-main}.} Since the argument is rather standard and follows a  similar path to the previous proof, we will omit some of the details.

To prove the result, one has to bound $\beta_N^*$ and $\alpha_N^*$ when $X=(\zeta_1,...,\zeta_n)$ and $\|\zeta\|_{L_\infty} \leq \kappa$. Using \eqref{eq:app-small-N} as above, it is evident that if $N \leq c_1n$,
\begin{equation*}
\E \sup_{f \in \cF_R \cap s \cD_{f^*}} \left|\frac{1}{\sqrt{N}}\sum_{i=1}^N \eps_i (f-f^*)(X_i) \right|  \leq  \gamma s \sqrt{N}
\end{equation*}
provided that
$$
s \leq c_2 \frac{R}{\sqrt{N}}\sqrt{\log(ec_1n/N)}
$$
for suitable constants $c_1$ and $c_2$ that depend only on $\kappa$ and $\gamma$. Moreover, using the notation of Theorem \ref{thm:main}, $\gamma=\tau Q_{\cF - \cF}(2 \tau)/16$ and thus, $\gamma$ is a constant that depends only on $\kappa$ as well.

Therefore,
\begin{equation*}
\beta_N^* \leq
\begin{cases}
\frac{c_3 R}{\sqrt{N}} \sqrt{\log\left(\frac{ec_1n}{N}\right)} &  \mbox{if} \ \ N \leq c_1n
\\
\\
0 & \mbox{if} \ \ N >  c_1 n.
\end{cases}
\end{equation*}
for constants $c_1$ and $c_3$ that depend only on $\kappa$.

Next, to bound $\alpha_N^*(\gamma,\delta)$, set $\sigma=\|W\|_{L_{2,1}}$, let
$$
V_s=\sup_{t \in 2RB_1^n \cap sB_2^n} \left|\frac{1}{\sqrt{N}} \sum_{i=1}^N W_i \inr{X_i,t} \right|
$$
and recall that $W$ and $X$ are independent. By a standard multiplier theorem (see, e.g., \cite{VW} Chapter 2.9),
$$
\E V_s \leq c_4 \sigma \max_{1 \leq m \leq N} \E \sup_{t \in 2RB_1^n \cap sB_2^n} \left|\frac{1}{\sqrt{m}} \sum_{i=1}^m \inr{X_i,t} \right|=(*),
$$
and since $\frac{1}{\sqrt{m}} \sum_{i=1}^m X_i$ is an isotropic, $c\kappa$-subgaussian vector, it follows from Talagrand's Majorizing Measures Theorem \cite{Tal:book} that
\begin{equation} \label{eq:multiplier}
(*) \leq c_5 \kappa \sigma \E \sup_{t \in 2RB_1^n \cap sB_2^n} \left|\sum_{i=1}^n g_i t_i \right|,
\end{equation}
where $g_1,...,g_n$ are independent, standard gaussian variables and $c_5$ is an absolute constant.

Applying Lemma \ref{lemma:d-sum} and recalling that $\gamma$ is a constant that depends only on $\kappa$, $\E V_s \leq (\gamma/4) \sqrt{N} s^2$ for the choice of
\begin{equation*}
s^2 =
\begin{cases}
c_6\frac{R\sigma}{\sqrt{N}} \sqrt{\log\left(\frac{ec_7n\sigma}{\sqrt{N}R}\right)} & \mbox{if } \ \  N \leq c_7n^2 \sigma^2/R^2,
\\
\\
c_6\frac{\sigma^2 n}{N} & \mbox{if} \ \ N >c_7n^2 \sigma^2/R^2.
\end{cases}
\end{equation*}
and constants $c_6,c_7$ that depend only on $\kappa$.

Finally, by Talagrand's concentration inequality for bounded empirical processes (see, e.g., \cite{BLM}) applied to the class $\{\inr{\cdot,t} : t \in 2RB_1^n \cap sB_2^n\}$ and relative to the probability measure endowed on $\R^n$ by $WX$, one has that with probability at least $1-\exp(-x)$,
$$
V_s  \leq 2\E V_s + c_8\left(\sigma s\sqrt{x} + \kappa^2 R \frac{x}{\sqrt{N}}\right)
$$
for an absolute constant $c_8$.

Indeed, for every $t \in 2RB_1^n \cap sB_2^n$ and since $W$ and $X$ are independent,
$$
\|W\inr{X,t}\|_{L_2} = \|W\|_{L_2} \|\inr{X,t}\|_{L_2} \leq 2\sigma s.
$$
Also,
$$
\|W\inr{X,t}\|_{L_\infty} \leq \|W\|_{L_\infty} \|\inr{X,t}\|_{L_\infty} \leq 2\kappa^2 R.
$$
Hence, for a constant $c_9$ that depends only on $\kappa$, $x=c_9 Ns^2 \min\{1/ R,1/\sigma^2\}$ and $\delta=\exp(-x)$, one has that with probability at least $1-\delta$, $V_s \leq \gamma \sqrt{N} s^2$. Therefore, $\alpha(\gamma,\delta) \leq s$, which completes the proof.
\endproof

\section{Concluding Remarks} \label{sec:concluding-rem}
Although it seems that the complexity parameters $\alpha_N^*$ and $\beta_N^*$ (and $k_N^*$ as well) depend on the unknown function $f^*$, in virtually all applications one may replace the indexing set $F \cap r \cD_{f^*}$ with $(\cF - \cF) \cap r\cD$, where $\cD$ is the unit ball in $L_2(\mu)$ -- with the obvious modifications to the definitions of the parameters. Moreover, if $\cF$ is centrally symmetric (i.e. if $f \in \cF$ then $-f \in \cF$) in addition to being convex, then $\cF - \cF \subset 2\cF$ and the indexing set becomes $2\cF \cap r \cD$.

\vskip0.5cm
The fact that $\beta_N^*$ depends on an average while $\alpha_N^*$ is based on a probability estimate is just an outcome of this presentation. It is possible to obtain a similar result using a `high probability' version of $\beta_N^*$, by using a slightly different argument. On the other hand, while one may replace $\alpha_N^*$ with an averaged version, the latter makes little sense in heavy-tailed situations, as passing from an average-based parameter to a high probability one reverts to the question of concentration. In heavy-tailed situations the mean need not represents the typical behaviour, and should not be used when trying to obtain very high probability bounds.

\vskip0.5cm

Another fact worth mentioning is that the results presented here can be extended well beyond the squared loss, to arbitrary convex loss functions (see \cite{Men-extend-ver}).

To explain why the choice of the squared loss is not essential for the method presented above, consider a smooth, increasing and even function $\ell$ that satisfies $\ell(0)=0$. The point-wise cost of predicting $f(X)$ instead of $Y$ is $\ell(f(X)-Y) \equiv \ell_f(X,Y)$. As above, set $f^*$ to be a minimizer in $\cF$ of the functional $\E \ell(f(X)-Y)$.
\begin{Assumption} \label{ass:general-case}
Assume that $f^*$ is unique, and, setting $\xi(X,Y)=f^*(X)-Y$, that $\E \ell^\prime(\xi)(f-f^*)(X) \geq 0$.
\end{Assumption}
Assumption \ref{ass:general-case} is not really restrictive. It is straightforward to verify that if $\ell$ is convex, the assumption holds when $\cF$ is closed and convex, or, for an arbitrary class $\cF$, when $Y=f_0(X)+W$ for some $f_0 \in \cF$ and an independent, mean-zero random variable $W$ that is independent of $X$.

The excess loss function associated with $f$ is $\cL_f =\ell_f - \ell_{f^*}$ and clearly an empirical minimizer of the loss is an empirical minimizer of the excess loss, and $P_N \cL_{\hat{f}} \leq 0$.

Given the data $(X_i,Y_i)_{i=1}^N$ and since $\E \ell^\prime(\xi)(f-f^*)(X) \geq 0$, a straightforward application of Taylor's expansion around each $\xi_i=f^*(X_i)-Y_i$ shows that for every $f \in \cF$
\begin{align} \label{eq:excess-general}
& P_N \cL_f =  \frac{1}{N}\sum_{i=1}^N \left( \ell\left( \left(f-f^*\right)(X_i)+\xi_i \right) - \ell(\xi_i) \right) \nonumber
\\
\geq &
\left(\frac{1}{N} \sum_{i=1}^N \ell^{\prime}(\xi_i) (f-f^*)(X_i) - \E \ell^{\prime}(\xi) (f-f^*) \right) + \frac{1}{2N} \sum_{i=1}^N \ell^{\prime \prime} (Z_i) (f-f^*)^2(X_i),
\end{align}
for midpoints $(Z_i)_{i=1}^N$ that belong to the intervals whose ends are $\xi_i$ and $\xi_i + (f-f^*)(X_i)$, and thus depend on $f$ and on the sample $(X_i,Y_i)_{i=1}^N$.

Just as in the squared-loss case, $f$ cannot be an empirical minimizer if $P_N \cL_f >0$. Therefore, using \eqref{eq:excess-general}, one may obtain a positive lower bound on $P_N \cL_f$ by identifying the levels $\bar{\alpha}_N$ and $\bar{\beta}_N$, for which, if $\|f-f^*\|_{L_2} \geq \bar{\beta}_N$ then
\begin{equation} \label{eq:beta-bar}
\frac{1}{2N} \sum_{i=1}^N \ell^{\prime \prime} (Z_i) (f-f^*)^2(X_i) \geq c \|f-f^*\|_{L_2}^2,
\end{equation}
for some constant $c$, and if $\|f-f^*\|_{L_2} \geq \bar{\alpha}_N$ then
\begin{equation} \label{eq:alpha-bar}
\left|\frac{1}{N} \sum_{i=1}^N \eps_i\ell^\prime(\xi_i)(f-f^*)(X_i)\right| \leq \frac{c}{4} \|f-f^*\|_{L_2}^2.
\end{equation}
On that event the estimation error satisfies
$$
\|\hat{f}-f^*\|_{L_2} \leq \max\{\bar{\alpha}_N,\bar{\beta}_N\}.
$$
Observe that for the squared loss, $\ell^\prime(\xi_i)=2 \xi_i$ and $\ell^{\prime \prime}(Z_i)=2$ regardless of $Z_i$. Hence, $\bar{\alpha}_N$ and $\bar{\beta}_N$ in the squared loss case lead to the parameters $\alpha_N^*$ and $\beta_N^*$ we have used here.

If $\ell$ happens to be strongly convex, e.g., if $\inf_{x \in \R} \ell^{\prime \prime}(x) \geq c_1>0$, the mid-points $Z_i$ need not play a real role in \eqref{eq:beta-bar}, as $\ell^{\prime \prime}(Z_i) \geq c_1$. Thus, the results presented here for the squared loss may be easily extended to a strongly convex loss. However, when $\ell$ is only convex the role of the midpoints becomes more significant and requires careful analysis. As a first step, one has to identify the correct level $\bar{\beta}_N$ for which, with high probability, the following holds: if $\|f-f^*\|_{L_2} \geq \bar{\beta}_N$, there is a subset of $\{1,...,N\}$ of cardinality that is proportional to $N$, and on which {\it both}
$$
\ell^{\prime \prime}(Z_i) \geq c_1 \ \ {\rm and} \ \ |f-f^*|(X_i) \geq c_2 \|f-f^*\|_{L_2}
$$
for constants that are independent of $f$.

\vskip0.5cm
A version of Theorem \ref{thm:main} for a general convex loss can be found in \cite{Men-extend-ver}.

\footnotesize {

\end{document}